# Automated Thermal Screening for COVID-19 using Machine Learning

Pratik Katte, Siva Teja Kakileti, Himanshu J. Madhu, and Geetha Manjunath

Niramai Health Analytix Pvt Ltd, Bengaluru, India

## Abstract

In the last two years, millions of lives have been lost due to COVID-19. Despite the vaccination programmes for a year, hospitalization rates and deaths are still high due to the new variants of COVID-19. Stringent guidelines and COVID-19 screening measures such as temperature check and mask check at all public places are helping reduce the spread of COVID-19. Visual inspections to ensure these screening measures can be taxing and erroneous. Automated inspection ensures an effective and accurate screening. Traditional approaches involve identification of faces and masks from visual camera images followed by extraction of temperature values from thermal imaging cameras. Use of visual imaging as a primary modality limits these applications only for good-lighting conditions. The use of thermal imaging alone for these screening measures makes the system invariant to illumination. However, lack of open source datasets is an issue to develop such systems. In this paper, we discuss our work on using machine learning over thermal video streams for face and mask detection and subsequent temperature screening in a passive non-invasive way that enables an effective automated COVID-19 screening method in public places. We open source our NTIC dataset that was used for training our models and was collected at 8 different locations. Our results show that the use of thermal imaging is as effective as visual imaging in the presence of high illumination. This performance stays the same for thermal images even under low-lighting conditions, whereas the performance with visual trained classifiers show more than 50% degradation.

## Introduction

In the last two years, Coronavirus and variants(COVID-19) have infected over 260 million people globally with approximately 5 million people losing their lives to this disease [1]. This makes it one of the deadliest pandemics the world has ever seen. COVID-19 is a communicable disease that spreads from an infected person to others through tiny particles called aerosols[2]. This contagious nature of this disease led to such staggering incidences in different countries across the globe. To combat the spread of the disease, stringent measures such as complete lockdowns were imposed till today in several parts of the world. Despite vaccination programmes, there are still challenges due to the new COVID-19 variants like Omicron, Delta etc. The World Health Organisation (WHO) strongly recommends wearing a mask and frequent fever screening in public places to reduce the spread of this disease [3]. Wearing masks in public places is an effective way to prevent the transmission of COVID-19 through respiratory droplets [4]. High body temperature (fever) is a common symptom of COVID-19, therefore a temperature check can help to identify potential people with COVID-19 [5]. Both these protocol measures are strictly followed in many countries, especially at crowded places such as malls, office spaces, airports, railway stations etc to limit the spread of the disease.

However, ensuring continued adherence to these protocol measures is challenging. Typically, visual inspection for masks and fore-head thermometers for temperature screening are employed in many countries. Both these screening measures would require the person to come in close contact with the other person. This might increase the risk of infection for the person who is screening as the research shows the potential of transmission of virus is upto 6 feet [6]. Infrared thermal imaging cameras and visual cameras are found to be a good alternative and have seen an increase in the adoption over the last year. Infrared thermal cameras measure temperature emitting from a surface from far and helps in detecting the temperature of persons walking into the premises without any close contact. On the other hand, either direct visual inspection or inspection through visual cameras are used to check whether the people are wearing masks.

Manual inspection can be tiresome and might often result in missing people either with high temperatures or without masks. In the literature, Artificial Intelligence (AI) has been used to automate these screening steps [7, 8, 9]. Most of these algorithms use the frames of the visual camera stream to do the major analysis (detection of faces and mask) and utilize the thermal cameras only to access the temperature values in the detected regions of interest. One major reason being the face detection task from visual images is well explored and the results show very high accuracies [10]. Along with this, there are a plethora of datasets to train complex deep learning architectures. However, the use of visual streams for analysis makes the performance highly dependent on lighting conditions. In the presence of high illumination, the visual characteristics are remarkable and the performance of algorithms are high. On the contrary, the visual characteristics are lost in the presence of low illumination and the performance of algorithms for screening would be low. This limits the use of the visual cameras for 24 hour screening and in places where maintaining high illumination is difficult. In addition, the use of visual cameras for screening has also drawn major concerns on privacy of the users [11] as the faces of persons are being saved without any proper consent.

A major benefit of thermal cameras is that they are invariant to lighting conditions as they capture the long wave radiation unlike the visual cameras that capture visual spectrum. Further, thermal cameras do not capture any visual image of the persons, thereby protecting the privacy of the visitors. This makes the use of thermal cameras a viable solution for 24-hour public surveillance monitoring. However, one big challenge in developing thermal based AI algorithms is the lack of public datasets to train the algorithms. There are hardly any open source datasets of thermal images of human faces with and without masks.

First main contribution of this paper is to open source our Niramai Thermal Imaging for COVID-19 (NTIC) dataset that consists of thermal images of persons walking into public premises like offices, malls and railway stations. Section 2 provides the details of this NTIC dataset. Second major contribution of this paper is to propose novel augmentations that improve the performance of deep learning algorithms for the tasks of fever detection and mask classification (Section 3). Section 4 and 5 discusses the results and compares our proposed thermal based techniques with visual based techniques for face detection and mask classification.

# 2. Dataset

In general, the availability of open source thermal datasets is very limited. In recent years, a few open source thermal datasets with human faces were published [12, 13]. However, these datasets consist of thermal head shots of different persons and are specifically captured for the purpose of face recognition. Tasks such as screening people for covid symptoms require thermal images of people entering into the premises. One of the main goals of this paper is to provide such a real-world field thermal image dataset called Niramai Thermal Imaging for COVID-19 (NTIC) to enable the infrared vision research community to develop deep learning algorithms for various problems. The following section describes three variants of the NTIC datasets:

## 2.1 NTIC Datasets

### 2.1.1 Thermal Surveillance Dataset

The thermal surveillance Dataset is obtained by capturing realistic scenarios of people entering into the hospitals, railway stations, offices and other public places. To collect this dataset, a recording setup consisting of a thermal camera and a laptop (illustrated in figure 1) is deployed in two locations. FLIR E75 is used for acquisition of thermal images. It has a standard Field of View (FOV) of 24° x 18°, a temperature sensitivity of less than 40 mK and an accuracy of ±2 °C. The camera is mounted on a tripod at a height of about 140cm placed at the entrance of the building. To save the thermal images, the thermal camera is connected to a laptop equipped with intel core i5-2450 and 8 GB ram. Each thermal image is of resolution 320x240 pixels and is saved in the laptop at a frame rate of 9Hz. The recording setup was functional throughout the whole day, i.e 24 hours, thereby allowing us to capture the thermal images at different lighting conditions/illuminations. Further, the dataset consists of images of people with masks as well as without masks. Figure 2 shows a few examples of thermal images from the thermal surveillance dataset.

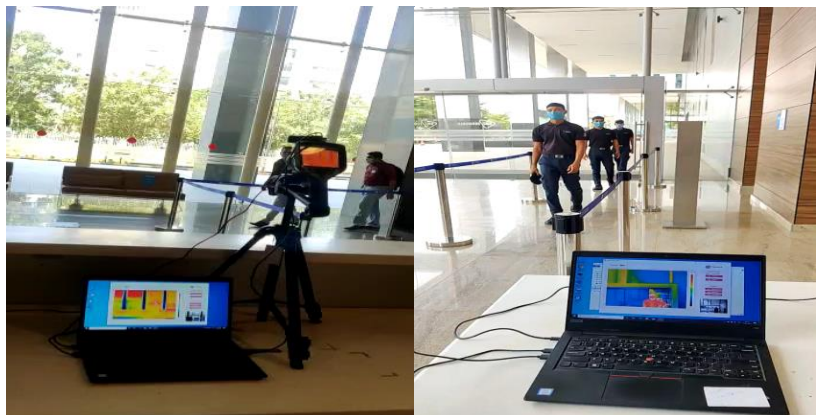

*Figure-1: Illustration of recording setup deployed in an office location.*

After analyzing 336 hours of videos with 9 frames per second, we manually segregated the frames with at least one person in the thermal image. This resulted in a dataset comprising 902 thermal images with 1354 people wearing masks and 213 people without masks. The numbers are slightly low as this dataset was collected during the first peak of COVID-19 when strict lockdowns were imposed in many parts of India. Each image in the dataset may consist of multiple persons in the frame with a minimum, mean and maximum of 1, 2 and 5 people, respectively, in the frame. Human experts are employed to identify the face bounding box of the persons and also to label the persons for mask classification in each of these frames. The details of annotation and the access to the dataset is discussed in Sec 2.2.

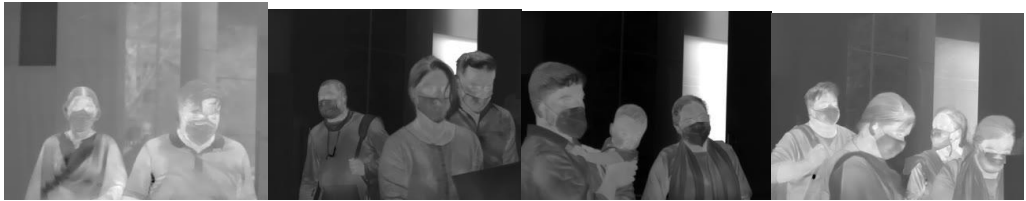

*Figure 2: Sample thermal images from the thermal surveillance dataset*

### 2.1.2 Augmented Surveillance Dataset

The size of the above discussed thermal surveillance dataset is still small when compared to other existing open source datasets. In our initial experimentations, we found that this small size of the dataset might lead to ineffective training of deep learning algorithms. To increase the size of the training dataset, we propose augmentations to the visual images to create a new augmented surveillance Dataset. Briefly, we considered visual images from an open source dataset and proposed transformations to convert these visual color images into gray thermal-lookalike images. Since there are many open sourced visual image datasets [14, 15], the transformations can result in a large augmented thermal dataset. The transformation consists of a two step rule as discussed below.

1) RGB to GrayScale: This involves conversion of a 3-channel RGB image to 1-channel grayscale image. The RGB pixels is converted into gray pixels using ITU-R BT.601 [16]:

$$Gray = 0.229R + 0.587G + 0.114B \quad (1)$$

Where R,G, and B are the pixel values of the red, green, and blue respectively.

2) Power Law Transformation: The database of open source visual images are captured in public places like malls and tech parks and the images comprise faces of persons with fair color complexion. The direct conversion of these images into grayscale results in bright face pixels. However, thermal pixels have a mixture of bright and dark pixels, which are minimally dependent on the color complexion. To make the objective image darker and more similar to actual infrared thermal images, gamma correction is applied for each of the images as described in equation 2.

$$O = I^{(1/G)} \quad (2)$$

Where I is the input image and G is the gamma value. Gamma values <1 will shift the image towards the darker end of the spectrum while gamma values > 1 will make the image appear lighter. A gamma value of G=1 will have no effect on the input image. A random gamma value in the range of 0.3 to 0.9 is applied to the visual images to darken the images.

In this paper, we considered the open source dataset provided in [15] to create this augmented surveillance dataset. After the transformation, we obtained 309 images with 543 persons. Out of 543 persons, 434 people are wearing masks and 109 people without masks. Figure 3 shows a sample input visual color image and its corresponding transformed gray thermal-like image.

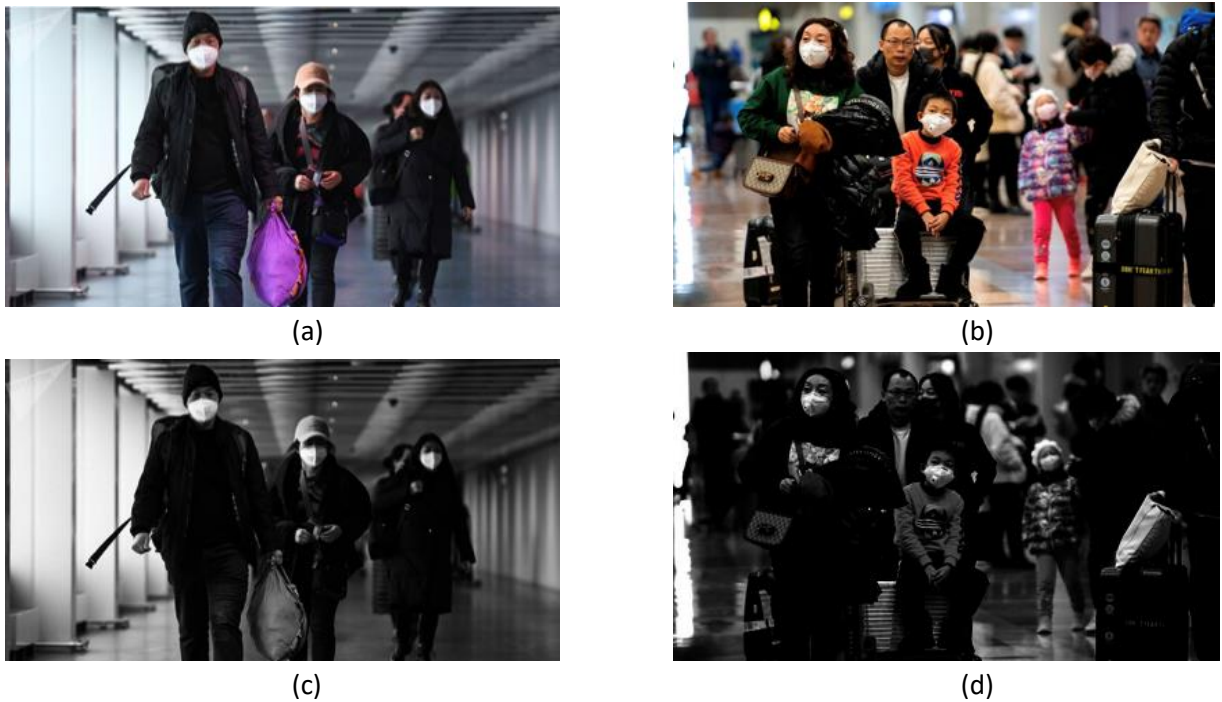

(a)  (b)
(c)  (d)

*Figure 3: (a) & (b) sample visual color images from the open source dataset [15]; (c) & (d) samples of transformed gray thermal-like images from augmented surveillance dataset.*

### 2.1.3 Lighting Dataset

The lighting dataset consists of both thermal images and their corresponding visual images of persons taken on different lighting conditions. The main aim of this dataset is to study and compare the variations of accuracy in face and mask detections with illumination for thermal and visual images. FLIR E75 that is used for the collection of thermal surveillance dataset is used for this data collection as well. FLIR E75 thermal camera automatically captures the corresponding visual image with a resolution 1280x960 pixels when a thermal image is captured. However, the visual and thermal images are not aligned due to the differences in FOVs of visual and thermal cameras. To align the images and to make the visual image resolution same as thermal image resolution, FLIR tools software provided by the FLIR Systems was employed.

We simulated four different lighting/illumination conditions (lux:0-25, lux:25-75, lux:75-150, lux:>150) by capturing thermal images at six different locations. These illumination intensities varied from completely dark spaces with 0 lux units to bright lighting spaces with illumination > 150 lux units. Lux meter was used to measure the illumination intensity in the location before each image capture. 25 participants volunteered for this image acquisition activity either in a single session or in multiple sessions. Each session comprised image capture of participants in the mentioned four different lighting conditions. For all the 25 participants, images were captured with and without the masks. Among the 25 participants, 14 are male and the remaining 11 are female persons. Further, 10 of these 25 participants appeared with and without eyeglasses as well. In total, there are 2*420 images in the dataset, where the factor 2 corresponds to thermal and visual images. Figure 4 demonstrates specimens from the dataset.

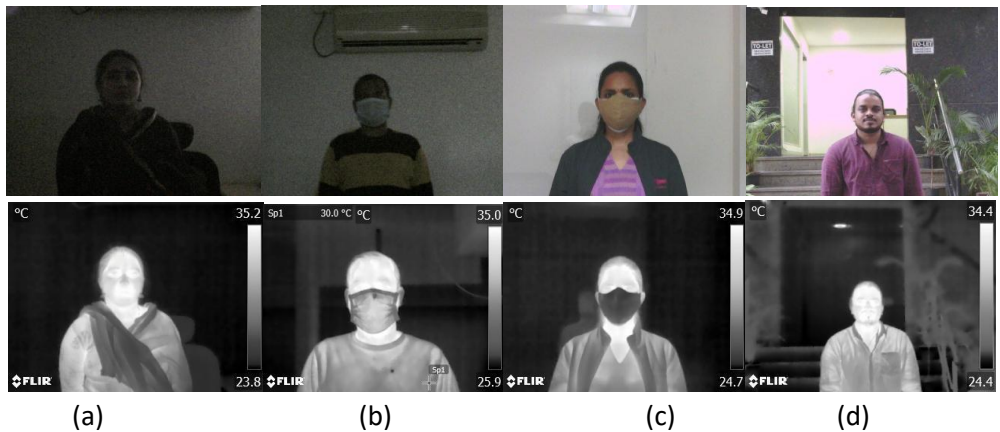

(a)          (b)          (c)          (d)

*Figure 4 - samples thermal and visual images of participants at different lighting conditions from lighting dataset(a) <25 lux, (b) 25-75 lux, (c) 75-150 lux, and (d) >150 lux.*

## 2.2 Annotation Description and Access to NTIC Datasets

All the three NTIC datasets are stored in google drive and the access to the drive link will be provided after submitting a data access request through https://www.niramai.com/contact page with subject as 'NTIC Dataset'. The drive link consists of 3 main folders corresponding to the dataset names. Each folder contains the thermal images in '.jpg' format and their corresponding bounding box and mask labels are saved in 'ground truth.txt'. The ground truth text file consists of image name, bounding box and a boolean label indicating mask. This ground truth is labeled and verified by at least two human experts.

## 2.3 Data Consent

Thermal surveillance dataset consists of only the thermal images of participants entering into our deployed sites and the visual information is neither open sourced nor saved with us. Since there is no direct way of correlating the thermal images to actual persons, we did not take explicit consent from the participants. However, the participants were aware and informed about the thermal image collection. The augmented surveillance dataset is a transformed version of an already existing open source dataset [15].

For the lighting dataset that comprised both visual and thermal images, we have explicitly informed the participants about the data collection and a written consent was collected.

# 3. Methodology

To summarize, most of the currently deployed COVID-19 fever screening systems use visual images for detection of face and mask and then use the corresponding face region in thermal images to obtain the temperature of the person. The use of thermal image alone for COVID-19 screening can provide advantages such as invariancy to the lighting condition and privacy aware screening. The existing state-of-the-art visual face detection algorithms [17, 18] are not directly applicable for thermal images as they result in poor Intersection Over Union (IOU) and precision in correctly detecting the faces, as detailed in Table 1.

| Pre-Trained RGB base Face Detection Models | Intersection Over Union (IOU) | Precision |
|---|---|---|
| Mobilenetv2-SSD | 52.23 | 4.38 |
| YOLOv3 | 24.9 | 30.61 |

Table 1 Performance metrics of state-of-the-art pretrained visual based classifiers when tested on the thermal surveillance test dataset.

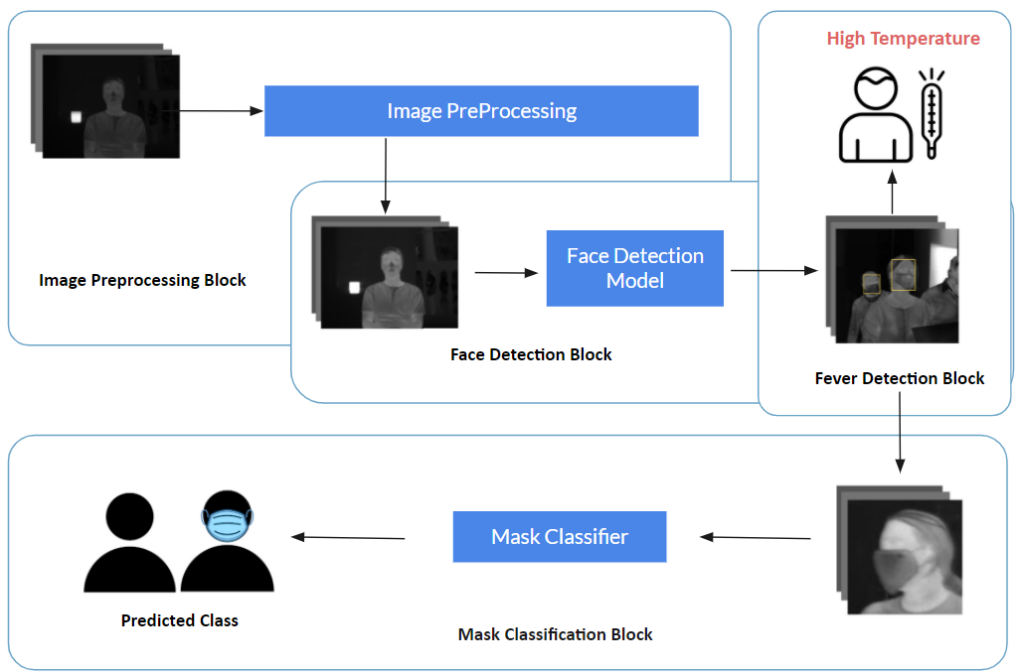

Figure 5: Block-diagram illustrating the end-to-end pipeline of automated COVID-19 screening with fever check and mask check.

The low performance is due to data drift from visual spectrum to long-wave infrared thermal spectrum. Therefore, there is a need for new algorithms to detect the faces and masks in the thermal images. However, lack of thermal dataset is a major challenge to train the deep learning algorithms. The proposed datasets can help in alleviating this problem. This section discusses the end-to-end pipeline (Figure 5) for detecting faces from thermal images followed by temperature detection and mask classification.

## 3.1 Image Preprocessing

Thermal images record the raw temperature values emitted from a body surface. The images in the thermal surveillance dataset consist of temperature values that can vary from 0℃ or lower to 100℃ or higher depending on the emitting surface. To convert these temperature values to [0, 1], a simple normalization technique involving the subtraction of the temperature matrix with minimum temperature value in the image followed by division with the temperature range was used initially. However, this led to loss of facial characteristics when there are extremely cold or hot objects in the scene as shown in Figure 6. Since the body temperature typically varies from 20°C to 40°C, a new temperature constrained normalization as shown in the equation 2 is proposed, where 'T' represents the temperature matrix and 'I' is normalized image with [0, 1] range.

$$I = \frac{T - max(min(T), 20)}{min(max(T), 45)} \qquad (3)$$

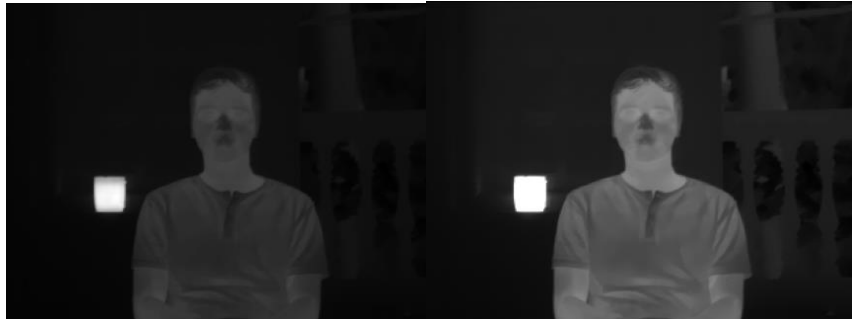

*Figure-6: 1): Normalised image before proposed image preprocessing. 2) Normalized image after proposed image preprocessing.*

## 3.2 Face Detection

Face detection is an important fundamental problem of computer vision and has been widely studied over the past decades. The use of deep learning algorithms have contributed significantly in improving the performance of detecting faces. However, much of the work proposed in the literature is applicable for the visual spectrum of images as discussed in the survey paper [10]. Two prominent light-weight architectures used for face detection in visual images are YOLOv3[17] and MobileNetv2-SSD [18, 19]. We have experimented with both the architectures and found that YOLOv3 is suitable for COVID-19 fever screening. Briefly, YOLOv3 is a 106 layer architecture with a combination of convolution and residual skip connections. It has 53 convolutional layers called Darknet-53 used for feature extraction from the input image. Each convolutional layer in the Darknet-53 is followed by batch normalization and Leaky ReLU activation. Other 53 layers are a combination of residual skip connections and upsampling layers stacked on top of the Darknet-53 for the detection task. In YOLOv3, the detection is done by applying 1x1 detection kernels on feature maps of three different sizes at three different places in the network. Using 9 anchor boxes in total, 3 for each scale, YOLOv3 uses logistic regression for predicting object confidence and class prediction and uses binary cross-entropy for calculating the classification loss for each class. To train the architecture, we use both thermal surveillance and augmented surveillance datasets as discussed in the experimentation section.

### 3.3 Fever Detection

To obtain the approximate body temperature from the face, different heuristic approaches were proposed in the literature [20]. These heuristics involved measuring maximum forehead temperature, maximum inner canthi temperature, maximum nasal temperature, maximum face temperature etc. In a recent evaluation study [21], it is found that full-face maximum temperature has high correlation with body temperature and provides a better discrimination capability for fever detection. Therefore, the maximum temperature in the detected face is used as a proxy for the body temperature. A participant is classified as high grade fever when this maximum face temperature is greater than 37.5℃.

### 3.4 Mask Classification

Mask classification on human faces is a recent requirement that came with COVID-19. The existing approaches in the literature use visual images for identifying if a person is wearing a mask or not [22,23, 24]. This classification task of masks from faces is a fairly simple problem as the mask might be seen as a dark patch of pixels in thermal images. Mask cloth has high absorption coefficient and weakens the infrared signals passing through it. This results in a dark patch of pixels compared to the rest of the facial temperatures. To make the inference faster, a lightweight architecture such as pre-trained mobileNetV2[18] is considered to classify if a person is wearing a mask or not. This pre-trained Mobilentv2 network with imagenet weights is slightly modified by adding dense, dropout and softmax layers to obtain the final classification output.

To train the architecture, both thermal surveillance dataset and augmented surveillance dataset are used. While using augmented surveillance dataset, the negative of the images are considered along with the transformed images for training the architectures. This is performed as the grayscale conversion of the color images with masks in general result in high pixel intensity (bright) for the mask region. As discussed, mask regions in thermal images have low temperatures due to the resistance provided by mask material. Therefore, the negative of images will convert these high pixel intensities (bright) to low pixel intensities (dark).

# 4. Experiments and Results

To evaluate the entire pipeline of this thermal based COVID-19 screening, we individually tested the performances of thermal face detection and mask classification deep learning architectures. We further compared these proposed thermal trained architectures with visual trained state-of-the-art techniques to measure the variation of their performances under different illumination conditions using the lighting dataset. To train the models for both the tasks, the initial 70% of thermal surveillance dataset and complete 100% of augmented dataset was used as a training dataset. While the remaining 20% and 10% of the thermal surveillance Dataset is used as validation and test datasets, respectively. This 70:20:10 split was performed according to the timestamps of the captured thermal images. This makes the last 10% images similar to a blind test set that is collected for validating the models. The detailed experimentation is as follows.

## 4.1 Face Detection

In the literature, architectures such as Faster-RCNN [25], RFCN [26], YOLOv3 and MobilenetV2-SSD are proposed for accurate detection of faces from visual images. We considered YOLOv3 and MobilenetV2-SSD architectures as they have high inference rate and can run on edge devices with low computational power. To evaluate the performance of thermal based face detection, both these architectures are trained on a training dataset comprising thermal surveillance and augmented surveillance datasets. During training, dataset samples were shuffled and a batch size of 8 samples was selected. Adam optimizer with $\beta = 0.9$, weight decay=0.95 was used to update the weights of the models. The learning rate was set to 1e-4 and is reduced by 10 times for every 5000 iterations. Performance metrics such as mean average precision (MAP) and mean Intersection Over Union (meanIOU) were used for comparison of different architectures.

Table 2 shows the results of YOLOV3 and Mobilenetv2-SSD with different combinations of the training set. As observed, the meanIOU is significantly increased by 4.6% for Mobilenetv2-SSD and 4.45% for YOLOv3 model when trained with the combination of thermal surveillance and augmented surveillance datasets. A slight improvement in MAP was also noticed when trained with the combination of both the datasets. To verify if these models are invariant to the illumination, we compared the visual trained classifiers and thermal trained classifiers on lighting dataset. Table 3 summarizes these results. Both visual trained classifiers performed better with a MAP > 90% only under good lighting conditions i.e illumination > 25 lux. This MAP is reduced to as low as 50% under lowest lighting conditions, i.e illumination <25 lux. These results show that the performance of visual trained classifiers is highly variable and requires a good lighting condition for better results. On the other hand, both the thermal trained classifiers resulted in 100% MAP in all the illumination conditions.

| Models | Training Dataset | meanIOU | MAP |
|---|---|---|---|
| Mobilenetv2-SSD | Thermal Surveillance Dataset | 76.7 | 88.9 |
| | Thermal Surveillance Dataset + Augmented Surveillance Dataset | 81.2 | 90 |
| YOLOv3 | Thermal Surveillance Dataset | 87.15 | 95.72 |
| | Thermal Surveillance Dataset + Augmented Surveillance Dataset | **91.60** | **96.48** |

Table 2: Results comparing YOLOv3 and MobileNetv2-SSD on the thermal surveillance test dataset.

| Lighting Conditions | YOLOv3 | | MobilenetV2-SSD | |
|---|---|---|---|---|
| | Visual trained | Thermal trained | Visual trained | Thermal trained |
| 0-25 lux | 54.60 | **100** | 50.36 | **100** |
| 25-75 lux | 94.68 | **100** | 90.34 | **100** |
| 75-150 lux | 98.91 | **100** | 92.22 | **100** |
| >150 lux | 98.72 | **100** | 84.72 | **100** |

Table 3: MAP results comparing the performance of visual trained and thermal trained classifiers on the lighting dataset.

## 4.2 Mask Classification

Transfer learning imparts the knowledge learned from the pre-trained classifier to a new classifier to aid its training. A small dataset can also achieve a good training effect through transfer learning. While pre-compiled visual based mask classification models were available on open source platforms [22,23, 24], there are no pre-compiled thermal based mask classification architectures due to lack of thermal datasets. To aid in the development of the mask classifier models, the thermal surveillance dataset was leveraged to manually curate and annotate a dataset of thermal face regions as shown in figure 7.

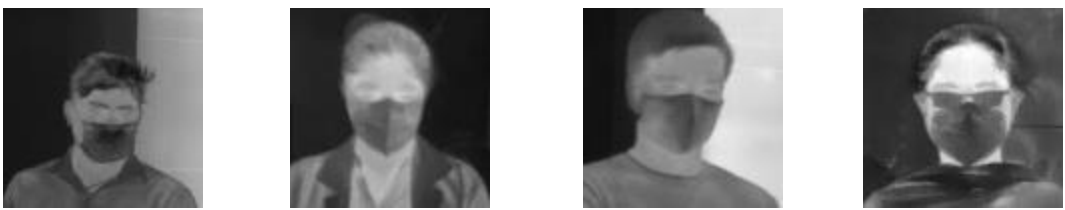

*Figure-7: Sample images of manually curated dataset for training mask classifier.*

As discussed in Sec 3.4, MobilenetV2 was considered for mask classification. MobilenetV2 consists of 155 layers including the top softmax classification layer with 1000 neurons corresponding to 1000 prediction classes. For the purpose of mask classification, we modified this architecture by replacing the softmax layer with a dense layer of 128 neurons followed by a dropout layer and then the softmax layer with two neurons. A transfer learning method was adopted for training this architecture where we used the imagenet weights as the default weights of the architecture and then re-trained with the thermal image training dataset. This allows the knowledge learned from the imagenet dataset to be transferred for the mask classification problem. Adam optimization with learning rate 1e-3 was used for updating the weights.

Table [4](#) summarizes the results obtained with the proposed architecture when trained with thermal surveillance dataset alone and with the combined thermal surveillance and augmented thermal datasets. The addition of this augmented dataset resulted in an overall 2% and 1% improvement in precision and recall, respectively. Further, the dense layer with 128 neurons produced a 9% and 7% improvement in precision and recall, respectively. These high results show that the mask detection can be performed directly from thermal imaging without the need for an external visual camera.

| Model | Precision | Recall |
|---|---|---|
| MobileNetV2 with 2 classes trained on thermal surveillance dataset. | 0.89 | 0.91 |
| Proposed model (mobileNetV2 + Dense+ softmax) trained only on thermal surveillance dataset | 0.98 | 0.98 |
| Proposed model (mobileNetV2 + Dense+ softmax) trained on combination of thermal surveillance and augmented thermal datasets | **1.00** | **0.99** |

*Table 4: Results comparing the precision and recall of different models on the thermal surveillance test dataset.*

In order to compare the variation in the results of the proposed thermal trained classifier with the recent visual trained state-of-the-art SSDMNV2 [24] classifier under different illuminations, we compared their performances on the lighting dataset. Table [5](#) shows their performances under the four considered illumination conditions. When SSDMNV2 model was used for inference on visual images from the lighting dataset, it performed better only on the images with good lighting conditions. In the presence of very low lighting conditions such as illumination < 25 lux, the mask classification precision was reduced to 64.4%. In contrast, the proposed thermal trained classifier did not vary significantly with illumination and resulted in a very high accuracy of 97% even when illumination is under 25 lux.

| Lightning Conditions | SSDMNV2[24] visual trained classifier | Proposed model with thermal training |
|---|---|---|
| 0-25 lux | 64.4 | **97** |
| 25-75 lux | 80.4 | **96** |
| 75-150 lux | 90.0 | **97** |
| >150 lux | 91.5 | **96** |

*Table 5- Results comparing the accuracy of* SSDMNV2[24] visual trained *mask classifier with proposed thermal trained classifier on lighting dataset.*

# 5. Discussion and Conclusion

With new variants like Omicron and Delta, the world is seeing new peaks with high incidences and deaths [27, 28]. Stringent prevention controls such as temperature checks and ensuring masks are worn by everyone could limit the spread of COVID-19. To implement these measures effectively, automated COVID-19 screening is a good alternative. This automated screening limits the need for a personnel to be at the site and ensures that the measures are followed properly. The choice of imaging is very important for the implementation of these COVID-19 protocol measures effectively. The traditional analysis on visual images might not be sufficient as they would require minimum illumination conditions for better performance. In this paper, we discussed the use of thermal imaging alone for ensuring both the COVID-19 protocol measures.

Thermal imaging involves the measure of long-wave infrared radiation and is non-invasive, portable and affordable. Due to these advantages, it is finding ways into various medical applications such as breast cancer detection, pain management, ruminative arthritis etc [29, 30, 31, 32] in recent years. When the COVID-19 hit the world, the demand for thermal cameras increased significantly to perform temperature screenings at airports, malls, restaurants etc. Automated temperature screening systems were also proposed for a safe and easy screening check [7, 8]. However, these automated screening systems used visual cameras to identify faces in the images and then extracted the temperature values of these identified faces from their corresponding thermal images.

On the contrary, the temperature values captured by the thermal imaging cameras are minimally affected by variations in the illumination intensity. This gives it an advantage over the traditional techniques that detect faces from visual images. As seen in tables. 3 and 5, the accuracy of models trained with thermal images is almost constant or does not vary significantly even when the illumination is < 25 lux. On the other hand, the accuracy is as low as 50% (random classification) when the illumination is < 25 lux for the models that are trained with visual images. In general, this low accuracy of models trained with visual images increased with increase in the illumination intensity with an exception for MobileNetV2-SSD face detection classifier that was found to result in slightly low accuracy when the lux >150. We believe that this low performance might be due to the architecture learning as the same was not observed with YOLOv3 under the same conditions.

Additionally, the performance of the thermal trained classifiers are found to be similar or sometimes better compared to visual trained classifiers. When the illumination > 150 lux, the face detection and mask classification models resulted in an average precision of 100% and 96%, respectively, for thermal trained classifiers and an MAP of 98.72% and 91.5%, respectively, for visual trained classifiers. Similar consistent results were obtained when we tested the thermal trained classifiers on the last 10% of the thermal surveillance dataset that resulted in an MAP of 96.8% and 100%, respectively, for face detection and mask classification.

These results with thermal images are very encouraging and show the potential of thermal imaging cameras in creating a high accuracy system for continuous COVID-19 monitoring that can work under any illumination. As thermal images capture temperatures and do not save any visual information of the persons, they can be deployed without worrying about the privacy of screening participants. The NTIC datasets used in this paper are open sourced under creative commons license to encourage academic and industrial communities in building robust thermal based systems that could help in ensuring the COVID-19 protocol measures in the pandemic. Further, the proposed augmentations discussed in this paper has helped in improving the classifiers performance by 2-4%. These augmented images are also open sourced as a part of NTIC datasets.

In conclusion, we discussed an end-to-end pipeline to create a system that could help in ensuring COVID-19 measures such as temperature check and mask check. The thermal based face detection and mask classification architectures along with the proposed augmentations resulted in better performance and are less variant to illumination when compared with the current state of the art architectures that are trained with visual images. We have not explicitly measured the accuracy of temperature values detected by thermal cameras, as the technique for body temperature approximation used in this paper is validated by Zhou et al. [21]. The future work involves adding more checks to the current system to reduce the false alarms due to normal elevated temperatures. Automated respiration rate from thermal images [33] is one way forward to achieve this.

# Acknowledgement

We like to acknowledge the authorities in the deployed sites for helping us with the data collection.

# Conflict of Interest

All the authors are employees of Niramai Health Analytix Private Ltd and have direct conflict of interest.